\pdfoutput=1

\documentclass[11pt]{article}

\usepackage{ACL2023}

\usepackage{times}
\usepackage{latexsym}

\usepackage[T1]{fontenc}

\usepackage[utf8]{inputenc}

\usepackage{microtype}

\usepackage{inconsolata}

\usepackage{times}
\usepackage{latexsym}
\usepackage{graphicx}
\usepackage{todonotes}
\usepackage{subcaption}
\usepackage{amsmath}
\usepackage{booktabs}
\usepackage{multirow}
\usepackage{todonotes}
\usepackage{wasysym}
\usepackage{tabularx}
\usepackage{pgfplots}
\usepackage{tikz}
\usepackage{amssymb}
\usepackage{pifont}

\usetikzlibrary{arrows.meta,decorations.pathmorphing,backgrounds,positioning,fit,petri}

\usepackage{enumitem}
\setlist[itemize]{align=parleft,left=0pt..1em}

\newcommand{\mypara}[1]{\noindent\textbf{#1}}

\title{Automatic QA Evaluator-Supervised Answer Generation}

\title{Answer Generation Models trained with Supervision from\\ Automatic Question Answering Evaluators}

\title{Learning Answer Generation using Supervision from \\Automatic Question Answering Evaluators}

\author{Matteo Gabburo$^{1}$\thanks{\ \ Work done as an intern at Amazon Alexa AI}\ , Siddhant Garg$^{2}$,  \textbf{Rik Koncel Kedziorski$^{3}$\thanks{\ \ Work completed at Amazon Alexa AI}\ , Alessandro Moschitti$^{2}$}\\
$^{1}$University of Trento , $^{2}$Amazon Alexa AI, $^{3}$Kensho Technologies, Inc. \\
\texttt{matteo.gabburo@unitn.it} \\ \texttt{\{sidgarg,amosch\}@amazon.com} \\
\texttt{rikka@kensho.com} \\
}

\begin{document}
\maketitle
\begin{abstract}
Recent studies show that sentence-level extractive QA, i.e., based on Answer Sentence Selection (AS2), is outperformed by Generation-based QA (GenQA) models, which generate answers using the top-$k$ answer sentences ranked by AS2 models (a la retrieval-augmented generation style). In this paper, we propose a novel training paradigm for GenQA using supervision from automatic QA evaluation models (GAVA). Specifically, we propose three strategies to transfer knowledge from these QA evaluation models to a GenQA model: (i) augmenting training data with answers generated by the GenQA model and labelled by GAVA (either statically, before training, or (ii) dynamically, at every training epoch); and (iii) using the GAVA score for weighting the generator loss during the learning of the GenQA model. We evaluate our proposed methods on two academic and one industrial dataset, obtaining a significant improvement in answering accuracy over the previous state of the art.
\end{abstract}

\section{Introduction}
\label{sec:introduction}

Recent research on retrieval-based Question Answering (QA) systems has been focused on two main tasks: (i) Answer Sentence Selection (AS2) e.g., \cite{garg2020tanda}, which, given a question and a list of answer candidates, chooses the most relevant answer that correctly answers the question; and (ii) Machine Reading (MR) e.g., \cite{Chen-Fisch-2017}, which, given a question and a reference text, involves finding a text span that directly answers the question. While effective, both the strategies (AS2 and MR) have some limitations: (i) the text might not include all the information necessary to answer a question, (ii) the text might include unnecessary, distracting information, or (iii) the text expresses the answer in a convoluted (indirect) format. Additionally, the text style and sentiment may be inappropriate for answering, or might be structurally too dependent on longer discourse context to enable usage as a stand-alone answer.

These drawbacks have motivated researchers to explore text generation systems for writing `better' answers in the open-domain abstractive QA setting. For example, in the MR domain, RAG~\cite{10.5555/3495724.3496517} generates an answer from a set of documents which are selected by dense passage retrieval models. For the domain of AS2, previous research has focused on summarizing answers from relevant paragraphs/evidences~\cite{lewis-etal-2020-bart}, or synthesizing information from the top ranked answer candidates of an AS2 system~\cite{hsu2021answer,muller-etal-2022-cross,gabburo2022}. 

The latter, termed as GenQA, has shown improvements in answer generation from the perspective of both answering accuracy and style suitability. The main distinguishing feature of GenQA from a generation-based MR approach is the length of the answer: the former uses an entire sentence as the target answer, while the latter in practice uses a short text (primarily targeting entity names). Therefore GenQA offers a more general and challenging research setting for answer generation.

Training effective GenQA models is made challenging by the cost and difficulty of obtaining large-scale, high quality training data. This typically requires human annotators to read the questions and relevant top $k$ retrieved paragraphs/sentences, and then re-write them into a self-contained, and concise natural answer (sentence/paragraph). 

Recent research~\cite{vu_ava_2021,bulian2022tomayto} has proposed effective automatic QA evaluation models based on transformer encoders for sentence-form answers. Training these QA evaluators only requires access to question answer pairs with annotations of correctness/incorrectness of the answers. This style of data annotation is much cheaper to perform than writing high-quality answers for training for GenQA models. In this work we explore the novel idea of using automatic QA evaluators for training GenQA models, which enables a faster and cheaper design implementation. 


In this paper, we reduce the amount of data needed for training a GenQA model using supervision from Automatic QA Evaluators.
Our first contribution is to propose GAVA: an automatic QA evaluation approach that extends AVA~\cite{vu_ava_2021} by (i) exploiting multiple reference answers and (ii) evaluating LM-generated answers instead of extracted answers. This way, we obtain a more robust and accurate QA evaluator that can effectively supervise the training of GenQA models. We propose three novel methods to use GAVA for refining the training of GenQA.

The first consists of (i) generating multiple possible answers using a baseline GenQA model for questions belonging to the GenQA training dataset, and (ii) then refining the set of generated answers by only retaining those with the highest GAVA scores (corresponding to ``correct" or ``high quality" answers). These generated answers are used as alternate gold standard answers (in addition to the annotators' written answers) to create additional training examples for GenQA. We term this approach GAVA-SDA (Static Data Augmentation).

The second approach extends GAVA-SDA, performing data augmentation dynamically at every epoch instead of off-line before training. This intuitively is more effective as the GenQA model continuously improves during the training. Specifically, at every epoch, we use GAVA to score the list of generated answers along with the $k$ input answer candidates. We then use the top scoring answer as the GenQA target and the next top-$k$ scoring answers as inputs for GenQA. We term this approach GAVA-DDA (Dynamic Data Augmentation).

The third approach uses GAVA as a scoring function for loss weighting during the training of GenQA. Specifically, we generate an answer using a GenQA model for a training sample, and weight the GenQA model loss of this instance using the GAVA score corresponding to the generated answer. Intuitively, this makes the GenQA model learn more from instances associated with higher GAVA-scoring answers (which corresponds to ``correct'' or ``high quality" answers). We term this approach GAVA-LW (Loss Weighting).

We perform empirical evaluation on two academic and one industrial QA dataset (de-identified customer questions from Alexa personal assistant), and show that our three proposed techniques using GAVA for training a GenQA model produce significant improvements in answering accuracy over a baseline GenQA approach. We also show that the answers generated by these improved GenQA models consistently achieve higher GAVA scores on average than the baseline. We will release the code along with the trained GenQA and GAVA models at \url{https://github.com/amazon-science/wqa-genqa-gava} to enable easy replication of our experimental results.


\section{Related Work}
\label{sec:relatedwork}

\mypara{Answer Generation:} Several research works~\cite{Izacard2021LeveragingPR,10.5555/3495724.3496517} have studied the problem of generating short answer spans (typically entity level) for MR systems. The most relevant of these works for GenQA is the work of ~\citet{DBLP:journals/corr/abs-2112-08688}, that trains an answer generation model using the evidentiality of retrieved passages. ~\citet{DBLP:journals/corr/abs-2110-06393} uses decoder cross-attention patterns to generate extractive answer spans. ~\citet{fajcik-etal-2021-r2-d2} generate answer spans by using a combination of a generative and extractive reader (aggregating information over multiple passages). An independent, but related line of research is question-based summarization, and there have been several research works in this field: ~\cite{Iida2019ExploitingBK,Deng2020JointLO}.

\citet{hsu2021answer} proposed the first formulation for generating complete answer sentences using evidences retrieved by an answer sentence selection (AS2) model. This model was termed GenQA, and it uses the top-$k$ most relevant answer sentence candidates for a question as input context to an encoder-decoder model to generate a natural sounding complete answer sentence for this question. ~\citet{muller-etal-2022-cross} extend GenQA for multiple languages by using answer sentence candidates from multiple languages as input context for the GenQA model. Recently, ~\citet{gabburo2022} propose training of GenQA models using unlabeled data by leveraging weak supervision from trained AS2 ranking models. This approach was shown to combine well with the supervised GenQA approach~\cite{hsu2021answer} to improve the answering accuracy. Note that all of these approaches are different from the ones described in the previous paragraph as they aim to generate complete answer sentences, and not just short answer spans. 

\mypara{Evaluation of QA Systems:} Token level similarity metrics like BLEU~\cite{papineni_bleu_2001}, ROUGE~\cite{lin-2004-rouge}, METEOR~\cite{banerjee-lavie-2005-meteor}, etc have been shown~\cite{reiter_structured_2018} to not extend to sentence-form QA evaluation. For MR tasks, \citet{yang-etal-2018-adaptations} adapt BLEU and ROUGE metrics, but limit their evaluation to only yes-no and entity questions. \citet{si-etal-2021-whats} uses multiple gold reference answers (extracted from Knowledge Bases) to be used as references for evaluating answer span extraction.

There have been several learnable automatic metrics: BERTScore \cite{zhang_bertscore_2020}, COMET \cite{rei_comet_2020}, BLEURT \cite{sellam_bleurt_2020},  etc. that have been proposed for some tasks in NLP such as MT and Summarization. These are based on transformer encoder models. \citet{chen-etal-2019-evaluating} proposed extending BERTScore for MR tasks using the question and paragraph context in addition to the answer. In similar line of work, \citet{vu_ava_2021} propose AVA which is an automatic QA evaluation metric that learns a transformer to encode the question, a reference gold answer and the target answer to be evaluated. Very recently, \citet{bulian2022tomayto} also present similar findings as AVA, by proposing BEM which can be used for evaluating sentence-level extractive QA (AS2). AVA and BEM have not been evaluated for GenQA systems previously. ~\citet{hsu2021answer} and \cite{gabburo2022} show that automatic metrics like BLEU, BLEURT, BERTScore do not correlate well with human judgements for evaluating accuracy of GenQA systems. We extend AVA for our experiments as the automatic QA evaluation system. 

\section{Automatic QA Evaluation using Multiple References (GAVA)}
\label{sec:ava}

~\citet{vu_ava_2021} propose AVA: an automatic evaluation models for QA based on a transformer encoder. It is applied to a question and a complete answer sentence to determine the correctness or incorrectness of the answer. Formally, we denote the AVA model with $\mathcal{E}$, which takes as input a question $q$, a target answer $a$, and a reference $r$, i.e., gold standard (GS) answer, and outputs a correctness probability score, $s \in [0,1]$. AVA is trained on the same labeled data of AS2, i.e., question answer pairs, where each question has multiple annotated answer candidates available.

Though the AVA approach was empirically shown to be accurate for evaluating AS2 systems~\citet{vu_ava_2021}, there are some limitations associated with it: (i) several questions may have diverse correct answers, e.g., "Tell me a winner of the US Open?", and (ii) the same answer may be expressed in very different formats, e.g., "How old is Joe Biden?", "Biden is 80 years old" v/s. "The president has just entered his life's eighth decade". Furthermore, AVA does not use negative references when evaluating correctness of answers, while incorrect answers can also help refine the prediction of correctness/incorrectness. Note that most AS2 datasets have multiple annotated answers (combination of correct and incorrect labels), and thus it is straightforward to use them for building data to train AVA with multiple positive and negative references. Intuitively, a GenQA system synthesizes an answer using different pieces of information spread across many relevant candidates (while suppressing any irrelevant information), aligning well with the idea of using multiple references for QA evaluation.

\begin{figure}[t]
\includegraphics[width=\linewidth]{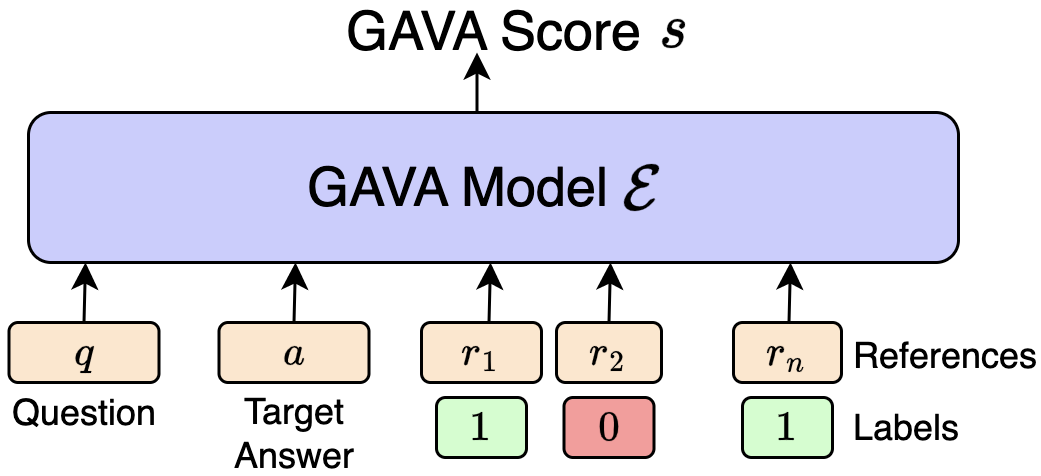}
\centering
\caption{Multi-reference GAVA that uses multiple positive and negative reference answers to evaluate the correctness/incorrectness of a target answer for a particular question and produces a score $s \in [0,1]$.}
\label{fig:ava}
\end{figure}

We term this approach: GAVA (AVA for generation-based models), which uses multiple references (combining positive and negative references) $\{r_1,r_2,\dots,r_n\}$.  Fig~\ref{fig:ava} shows the GAVA architecture: which uses a transformer encoder, taking as input: a question $q$, a target answer $a$, and $n$ references. The information about the nature of the positive/negative references is encoded by prepending each reference with a prompt indicating the type of reference it is.


\subsection{Comparison between GAVA and AVA}
\label{sec:gavavsava}

In subsequent sections of the paper, we will use the QA evaluator as a teacher to transfer knowledge for training GenQA models. We hypothesize that this knowledge transfer improves the answer generation capability of GenQA models by enabling the model to discriminate between good and poor supporting answer candidates. Thus a stronger QA evaluator teacher will benefit in training more effective GenQA models. Here we perform an empirical comparison between GAVA and the baseline AVA model, to show that the former achieves a higher correlation with human annotations. 

We consider two Answer Sentence Selection (AS2) datasets: WikiQA~\cite{yang2015wikiqa} and TREC-QA~\cite{wang-etal-2007-jeopardy}. We use a DebertaV3-Large~\cite{he2021debertav3} pre-trained model for both AVA and GAVA, and set $n{=}5$ reference answers per question for the latter. We measure the Pearson correlation between the human annotations and the two QA evaluators for each dataset under two configurations in Table~\ref{tab:comparisonavagava}: (i) \textbf{Extractive QA (AS2)} using the answer candidates available in the datasets, and (ii) \textbf{Generative QA (GenQA)} using answers that are written using a T5-Large GenQA model~\citet{hsu2021answer}. The results indicate the empirical superiority of GAVA over AVA as an automatic QA evaluation metric, which stems from the usage of multiple references, including negative ones.

\begin{table}[h]
\small
\resizebox{\linewidth}{!}{
\begin{tabular}{@{}lccccc@{}}
\toprule
\multirow{2}{*}{\textbf{Model}} & \multicolumn{2}{c}{\textbf{Extractive QA (AS2)}} & & \multicolumn{2}{c}{\textbf{Generative QA (GenQA)}} \\
\cmidrule{2-3} \cmidrule{5-6}
                                & \textbf{WikiQA}      & \textbf{TREC-QA}     & &\textbf{WikiQA}    & \textbf{TREC-QA}   \\ \midrule
\textbf{AVA}                    & 0.632                & 0.797        &       & 0.678              & 0.647             \\
\textbf{GAVA}                   & \textbf{0.676}       & \textbf{0.842}     & & \textbf{0.690}     & \textbf{0.671}    \\ \bottomrule
\end{tabular}}
\caption{Comparison between AVA and GAVA on WikiQA and TREC-QA. The models are compared in terms of Pearson correlation between the evaluation system prediction and the human evaluation. The best results for each dataset are highlighted in bold.}
\label{tab:comparisonavagava}
\end{table}

\section{Generative QA (GenQA)}
\label{sec:genqa}
\label{ssec:problem_setting}

Answer generation-based QA (GenQA) refers to a text generation model for generating an answer to a question. Specifically, a generation model $\mathcal{M}$ is provided a question $q$ and some context as the input, and generates an answer $g$. ~\citet{hsu2021answer} proposed GenQA for generating natural sounding complete answer sentences by leveraging labeled datasets having high quality human authored answers as the targets for generation. 

Specifically, a dataset, $\mathcal{D}$, for training a GenQA model, $\mathcal{M}$, contains examples of the format: $\Big(q, \{a_1,a_2,\dots,a_k\},t\Big)$ where $q$ is the question,  $\{a_1,\dots,a_k\}$ are the $k$ answer candidates used as input context to $\mathcal{M}$, and $t$ is the target output answer (GS human authored answer).

~\citet{gabburo2022} extended this line of work by proposing a novel approach to train GenQA models using unlabeled data by transferring knowledge from an AS2 model (that is used to produce silver labels). Specifically, for each question $q$, the AS2 model is used to rank a set of answer candidates without having any label of correctness/incorrectness for answering the question. The top ranked answer is used as the generation target for the GenQA model, while the question along with the next $k$ top-ranked answers are used as the input for the GenQA model.

\section{GAVA for Training GenQA}
\label{sec:methodology}
In this section, we propose three approaches for training GenQA models using GAVA.

\subsection{Static Data Augmentation ({\small GAVA-SDA})}
\label{ssec:ava_sda}

We create new training examples starting from $\mathcal{D}$, using a GAVA model, $\mathcal{E}$, and a base GenQA model $\mathcal{M}_0$ (trained only on $\mathcal{D}$). The new examples are added to $\mathcal{D}$ to create an improved training corpora for learning an improved GenQA model, $\mathcal{M}$. 

\begin{figure}[t]
\includegraphics[width=\linewidth]{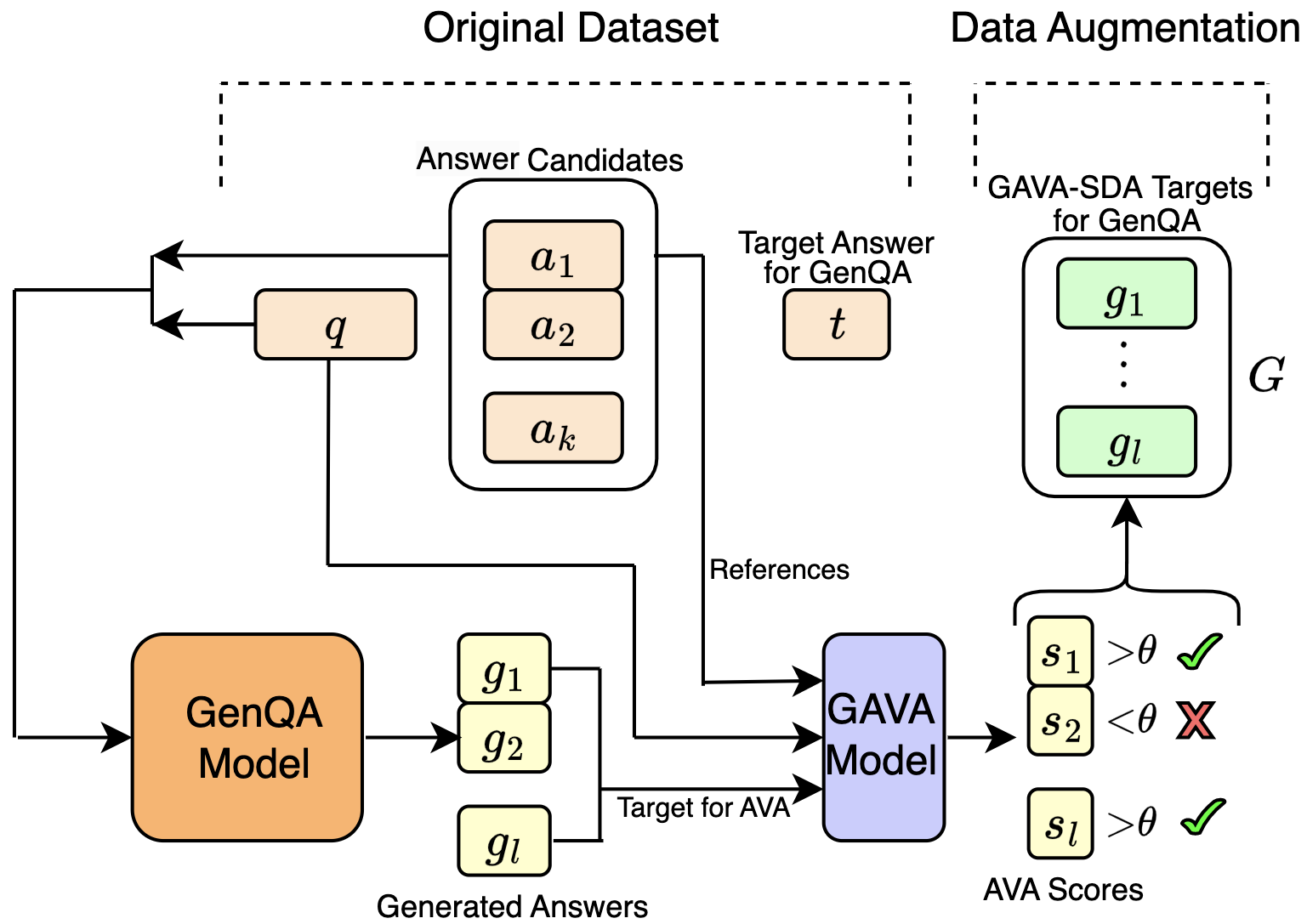}
\centering
\caption{Illustrative representation of the GAVA-Static Data Augmentation (GAVA-SDA) approach. }
\label{fig:ava_sda}
\end{figure} 

For every question, $q {\in} \mathcal{D}$, along with its answer candidates $\{a_1,\dots,a_k\}$ as input context, we apply $\mathcal{M}_0$ to generate multiple possible answers $\{g_1,g_2,\dots,g_l\}$,  using a probabilistic decoding approach~\cite{10.1162/tacl_a_00502}. Then, we apply the GAVA model to each of the generated answers $g_i$ to obtain GAVA scores, $s_i$, of correctness i.e.,  $\mathcal{E}(q,g_i,\{a_1,..,a_k\})$. 
Then using a pre-defined threshold $\theta$, we filter and pick only those answers $G=\{g_i: s_i \ge \theta\}$. 
We use this set of generated and filtered answers as alternate targets for generation to produce new examples for training GenQA, $\Big(q, \{a_1,\dots,a_k\},g\Big)$, where $\Big(q, \{a_1,\dots,a_k\},t\Big) {\in} \mathcal{D}$ and $g{\in} G$. Fig.~\ref{fig:ava_sda} illustrates this approach.

It should be noted that:
(i) $\theta$ is a parameter that can be tuned to increase the probability of correctness of the generated answers $g_i$. However, a very high $\theta$ will lead to filtering out a majority of the generated answers, leading to a very small augmented set (trade-off between size and quality). For our experiments, we used a value of $\theta$ that generated a large set of good quality diverse answers, as indicated by the GAVA score. 
(ii) Training a GenQA model on the augmented data can refine its predictions, biasing the generation towards ``good-quality" answers. Overall, this produces improvement in quality and accuracy of the generated answers.


\subsection{Dynamic Data Augmentation ({\small GAVA-DDA})}
\label{ssec:ava_dda}

We can improve the GAVA-SDA approach by producing new examples at regular intervals during the training, e.g., at the beginning of every epoch. This makes the data augmentation approach more adaptive, improving the learning of the GenQA model $\mathcal{M}$. 
As $\mathcal{M}$ improves during training, it will generate improved and higher quality answers, which can then be selected by GAVA to augment for the subsequent iterations. These `higher-quality' answers can, in turn,  improve the GenQA model's generation ability. In other words, instead of using a static base GenQA model $\mathcal{M}_0$, for the generation of the augmented data, we use the latest GenQA model $\mathcal{M}$, trained on the latest augmented data in the training routine. 

Additionally, we refine the input context to the GenQA model during training.  After obtaining the filtered set of generated and selected answers from GAVA: $G$, we combine them with the  answers from $D$, i.e., $\mathbb{A}=\{a_1,\dots,a_k, t\} \cup G$. We then use $\mathcal{E}$ to score $\mathbb{A}$, and pick the topmost ranked answer as the target for GenQA, and the next $k$ top ranked answers as the input context for GenQA (following the same idea as \cite{gabburo2022}). Intuitively, this can improve the quality of both the input context that the GenQA model is using for generation, as well as the output target answer.

We combine the above two modifications into a single approach and call this Dynamic Data Augmentation (GAVA-DDA).

\subsection{Loss Weighting ({\small GAVA-LW})}
\label{ssec:avaintheloop}

GAVA-SDA and GAVA-DDA transfer the knowledge of the GAVA evaluation model for training GenQA by augmenting training data. Both approaches do not modify the GenQA training approach. In contrast, GAVA-LW uses the GAVA score to modify the GenQA training loss.

\begin{figure}[t]
\includegraphics[width=\linewidth]{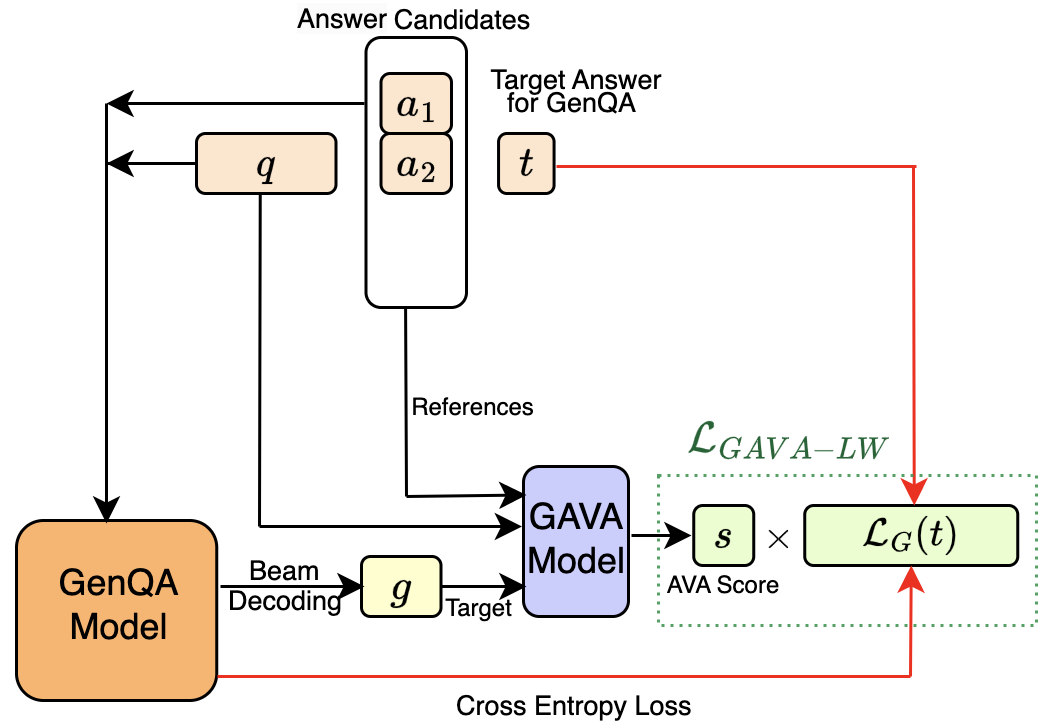}
\centering
\caption{Illustrative representation of the GAVA-Loss Weighting (GAVA-LW) approach. }
\label{fig:ava_lw}
\end{figure}

More formally, for training $\mathcal{M}$ with an example $\Big(q, \{a_1,\dots,a_k\},t\Big) \in \mathcal{D}$, we apply three steps: (i) compute the standard cross entropy loss $\mathcal{L_G}(t)$ of GenQA model $\mathcal{M}$ on input $\Big(q, \{a_1,\dots,a_k\}\Big)$ with target $t$ (ii) run inference procedure on $\Big(q, \{a_1,\dots,a_k\}\Big)$ to obtain model generation $g$ 
 (iii) compute the GAVA score of $g$ using $\mathcal{E}(q,g,\{a_1,..,a_k\})$. We then use the GAVA score to weight the GenQA training loss as follows:
\begin{equation*}
    \mathcal{L}_{GAVA-LW} {=} \Big(1-\mathcal{E}\big(q,g,\{a_1,..,a_k\}\big)\Big) {\times} \mathcal{L_G}(t)
\end{equation*}
The GAVA-LW approach is illustrated in Fig.\ref{fig:ava_lw}. Intuitively, we want to make the model learn to improve its predictions for examples where the answer quality given by GAVA is low. Thus, for these examples, we give a weight to the training loss with the GAVA score. 
The $\mathcal{L}_{GAVA-LW}$ formulation (i) helps the model emphasize \emph{harder} training samples (on which the model is currently not performing well) during learning, and (ii) trains a stronger more generalized system. 

\section{Experiments}
\label{sec:datasets_models}

\subsection{Datasets}
\label{ssec:datasets}

For training and evaluating our models, we consider two academic and one industrial dataset representing real world customer questions.

\mypara{WQA} Web Question Answers (WQA) is a public dataset defined in ~\cite{zhang-etal-2021-joint}. The dataset contains 149,513 questions, each associated with about 15 answer candidates. Both questions and answers are retrieved from a large-scale web index. Each QA pair is manually annotated for answer correctness by professional annotators. 

\mypara{MS-MARCO} \citet{bajaj2018ms} proposed MS-MARCO, originally for MR tasks, comprising $\sim$800k queries retrieved from the Bing search engine along with  $\sim$10 labeled answer passages. Following ~\citet{gabburo2022}, we transform MS-MARCO to obtain a large dataset of QA pairs, where the answers are sentences and not passages/paragraphs. Using a SOTA AS2 model (DeBERTav3-xl~\cite{he2021debertav3} trained on the ASNQ ~\cite{garg2020tanda} dataset), we pick the top-2 ranked answer sentences from a positively labeled passage as positive answer candidates for the question. Similar to ~\citet{gabburo2022}, we randomly sub-sample 1k questions from the dev.~set for evaluation (we use human annotations for our experiments and using the entire 100k dev.~set would be extremely expensive to annotate).

\mypara{IQAD} Industrial QA Dataset~\cite{garg-moschitti-2021-will,diliello2022pretraining} is a large scale internal industrial QA dataset containing non-representative de-identified user questions from Alexa personal assistant. IQAD contains ${\sim}$10k questions, each having ${\sim}$200 answer candidates retrieved using a large scale web index (over 100M documents). Each question has $\sim$15 answer candidates with human annotations of correctness/incorrectness. Results on IQAD are presented relative to a baseline due to the data being internal.

\begin{table*}[h!]
\centering
\resizebox{\linewidth}{!}{
\begin{tabular}{@{}lccccccccc@{}}
\toprule
  \multirow{2}{*}{\textbf{Approach}} & &
  \multicolumn{2}{c}{\textbf{WQA}} &&
  \multicolumn{2}{c}{\textbf{MS-MARCO}} &&
  \multicolumn{2}{c}{\textbf{IQAD}} \\
   &&
  \multicolumn{1}{c}{\textbf{Accuracy}} &
  \textbf{GAVA-Score} &&
  \multicolumn{1}{c}{\textbf{Accuracy}} &
  \multicolumn{1}{c}{\textbf{GAVA-Score}} &&
  \multicolumn{1}{c}{\textbf{Accuracy}} &
  \multicolumn{1}{c}{\textbf{GAVA-Score}} \\ \midrule
 GenQA-WS~\cite{gabburo2022}      && 0.655     & 0.409 && 0.775 & 0.770 &&   Baseline   & Baseline  \\
\midrule
 \textbf{(Ours)} GAVA-SDA && \textbf{0.868}     & 0.498 && \textbf{0.877} & \textbf{0.869} &&  +8.55\%    &  -1.48\%  \\
 \textbf{(Ours)} GAVA-DDA && 0.769     & 0.439 && 0.843 & 0.855 &&  \textbf{+9.85\%}    &  \textbf{+0.54\%}  \\ 
 \textbf{(Ours)} GAVA-LW  && 0.796     & \textbf{0.527} && 0.794 & 0.784 &&  +8.81\%    &  +0.51\%  \\\bottomrule
\end{tabular}
}
\caption{Answering accuracy (manual evaluation) and AVA-Score on WQA, MS-MARCO and IQAD datasets. Results on IQAD are presented relative to the baseline (due to the data being internal). For WQA and MS-MARCO, we use an AVA model trained on WQA, and for IQAD we use an AVA model trained on IQAD. Best results for each dataset are highlighted in bold.}
\label{tab:bigtable}
\end{table*}

\subsection{Models and Evaluation}
\label{ssec:models}

For our experiments we consider two types of models (i) GAVA evaluation models, as described in Section~\ref{sec:ava}, and (ii) GenQA models, using techniques described in Section~\ref{sec:methodology}. For GAVA $\mathcal{E}$, we use a DebertaV3-Large~\cite{he2021debertav3} pre-trained model using up to $n{=}5$ reference answers per question. We train two GAVA models: one on WQA and one on IQAD, using the former for both the public datasets.~\footnote{WQA contains human annotations of answer correctness which can be used as references for training a strong GAVA model. The answer sentence version of MS-MARCO does not contain human annotations of answer correctness.} For GenQA, we consider a baseline model from \citet{gabburo2022}, which is a T5-Large~\cite{raffeletall2020} encoder-decoder transformer trained using weak supervision on MS-MARCO. We consider this as the baseline GenQA model $\mathcal{M}_0$, and apply all of our techniques: GAVA-SDA, GAVA-DDA, GAVA-LW starting from this. Unless otherwise stated, we use $\theta{=}0.9$ for GAVA-SDA and GAVA-DDA. For the GenQA models, we take $k{=}5$ answer candidates as inputs, and select the best checkpoint, corresponding to highest AVA-Score on the development set. We present complete experimental details in Appendix. 

We perform human evaluation of the generated answers using
Amazon MTurk (5 annotations per QA pair, taking average of these scores). We selected a pool of turkers having an approval rate higher than $95$\% with more than $500$ approved hits. We designed our annotation task by providing the annotator with (i) the question, (ii) the generated answer, and (iii) a correct reference answer. For each hit (question + generated answer pair),  we pay the turker $0.1$\$ and obtain $5$ independent annotations. Using these annotations, we compute the answering accuracy over the entire evaluation set: number of correct answers divided by the total number of generated answers. We also evaluate models using the automatic GAVA metric.

\subsection{Main Results}
\label{ssec:quantitativeresults}
We evaluate GenQA models trained with our three proposed techniques in Table~\ref{tab:bigtable} using human evaluation of accuracy and GAVA-Score (automatic evaluation). We observe that across all datasets, our approaches outperform the baseline and are able to improve GenQA training, as evidenced by both human and automatic evaluation.  

Specifically for WQA, we observe that the GAVA-SDA approach achieves the highest answering accuracy (improving an impressive +21.3\% points over the baseline). The experiments on WQA indicate the ideal case, where we can have a GAVA model trained on the same dataset (due to availability of some annotations of correctness). We even observe improvement in the GAVA score for our approaches (which is expected, since we are using this model to supervise the training of GenQA). Interestingly, we do not see a perfect correlation between the human-induced and GAVA-induced relative ordering of the four techniques.

On MS-MARCO, we again observe that GAVA-SDA achieves the highest answering accuracy (+10.2\% points over the baseline), and here there is a perfect correlation between the human evaluation and GAVA. This evaluation on MS-MARCO demonstrates the transferability of using GAVA for teaching GenQA across data distributions (the GAVA model used here is trained on WQA, as the sentence version of MS-MARCO does not have human annotations).

On the industrial dataset, IQAD, we observe that the GAVA-LW loss weighting approach achieves the highest accuracy (+9.85\% relative improvement over the baseline). The results on IQAD lend support to our approaches extending to a real world scenario with actual customer questions.

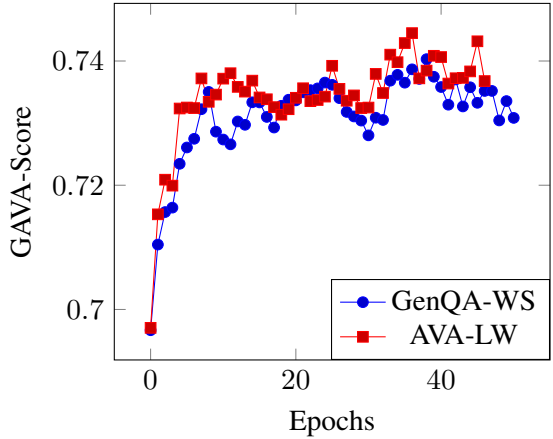
\begin{figure}[t]
    \begin{tikzpicture}
    \begin{axis} [
        width=0.95\linewidth, 
        xlabel=Epochs, 
        ylabel=GAVA-Score, 
        legend style={at={(1,0)},anchor=south east}
    ]
    \addplot table [x=epoch,y=baselineavascore, col sep=comma] {data.csv};
    \addplot table [x=epoch,y=avaweightavascore,col sep=comma] {data.csv};
    \legend {GenQA-WS,AVA-LW};
    \end{axis}
    \end{tikzpicture}
    \vspace{-0.5em}
\caption{Comparison of baseline GenQA-WS and AVA-LW on WQA in terms of GAVA-Score (on validation split) varying across training epochs (GAVA-LW achieves higher GAVA-Score throughout training).}
\label{fig:ablation_avascore_ws_vs_avalw}
\end{figure}

\begin{table*}[t]
\centering
\resizebox{0.85\linewidth}{!}{
\begin{tabular}{@{}lccccc@{}}
\toprule
\textbf{Approach}    & \textbf{Manual}    & \textbf{GAVA-Score} & \textbf{BLEURT} & \textbf{BERTScore} & \textbf{BLEU} \\ \midrule
GenQA-WS       & 0.775 & 0.770 & 0.587 & 0.492 & 30.8 \\
GAVA-SDA($\theta{=}0.5$) & 0.846 & 0.857 & 0.578 & 0.496 & 35.2 \\
GAVA-SDA($\theta{=}0.7$) & 0.861 & 0.864 & 0.576 & 0.491 & 34.9 \\
GAVA-SDA($\theta{=}0.9$) & 0.877 & 0.869 & 0.573 & 0.486 & 34.1 \\
GAVA-DDA      & 0.843 & 0.855 & 0.627 & 0.541 & 38.1 \\
GAVA-LW       & 0.794 & 0.784 & 0.627 & 0.502 & 32.2 \\ \midrule
\multicolumn{2}{c}{Correlation (Pearson)} & 0.979  & -0.429 & -0.035 & 0.679  \\
\multicolumn{2}{c}{Correlation (Spearman's)}   & 1      & -0.812 & -0.6   & 0.429  \\ \bottomrule
\end{tabular}
}
\caption{Evaluation of different GenQA models using automatic evaluation metrics: BLEU, BLEURT, BERTScore in addition to GAVA-Score on the MS-MARCO dataset. We present the correlation each metric achieves with human annotation. GAVA achieves the best correlation with human evaluation of answer accuracy.}
\label{tab:ablation_correlation}
\end{table*}

\subsection{Analysis and Ablations}
\label{ssec:ablations}

\mypara{Variation of GAVA Score over Training} To understand how the GAVA score of our proposed techniques improves over the baseline, we plot its variation over the training epochs. We pick the MS-MARCO dataset and GAVA-LW as our approach to compare with, and present results in Fig.~\ref{fig:ablation_avascore_ws_vs_avalw}. From the figure, we observe that the GAVA-LW achieves a higher GAVA score than the GenQA baseline throughout the training regime. This demonstrates the knowledge transfer from the GAVA model for training GenQA, as the GenQA model is able to achieve a higher GAVA score over training epochs.

\begin{table}[h]
\centering
\resizebox{\linewidth}{!}{
\begin{tabular}{@{}cccc@{}}
\toprule
\textbf{$\boldsymbol{\theta}$} & \textbf{| Augmented Set |} & \multicolumn{1}{l}{\textbf{Accuracy}} & \multicolumn{1}{l}{\textbf{GAVA-Score}}  \\ \midrule
0.5 &  91,348  & 0.846  & 0.857 \\
0.7 &  84,272  & 0.861  & 0.864 \\
0.9 &  69,557  & \textbf{0.877}  & \textbf{0.869} \\ \bottomrule
\end{tabular}
}
\caption{Variation of GenQA accuracy by changing $\theta$ for GAVA-SDA approach, on the MS-MARCO dataset. We present human and automatic (GAVA-Score) evaluation. | Augmented Set | indicates the number of data augmentation examples created using a particular value of $\theta$ (Lower $\theta\rightarrow$ more augmentation examples).}
\label{tab:results_thresdholds_ablation}
\vspace{0.2em}
\end{table}

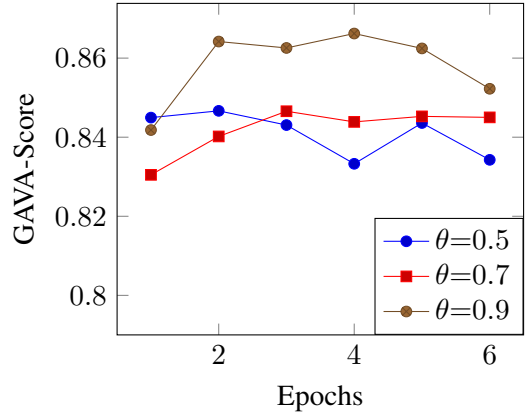
\begin{figure}
\begin{tikzpicture}
\begin{axis} [
    width=0.9\linewidth, 
    ymin=0.79,
    xlabel=Epochs, 
    ylabel=GAVA-Score, 
    legend style={at={(1,0)},
    anchor=south east}
    ]
    \addplot table [x=epoch,y=th05,col sep=comma] {ablationthresh.csv};
    \addplot table [x=epoch,y=th07,col sep=comma] {ablationthresh.csv};
    \addplot table [x=epoch,y=th09,col sep=comma] {ablationthresh.csv};
    \legend {$\theta{=}0.5$,$\theta{=}0.7$,$\theta{=}0.9$};
\end{axis}
\end{tikzpicture}
\vspace{-0.8em}
\caption{Comparison of three different GAVA-SDA models trained on MS-MARCO using different thresholds $\theta$ in terms of GAVA-Score (on validation split) varying across training epochs. The model trained with the largest value of $\theta$ achieves the highest GAVA-Score.}
\label{fig:ablation_thresholds}
\vspace{-0.3em}
\end{figure}

\mypara{Variation of Threshold $\boldsymbol{\theta}$ for GAVA-SDA:} As discussed in Section~\ref{sec:methodology}, $\theta$ is a tunable parameter that decides the quantity v/s quality trade-off for data augmentation. We aim to study how the choice of $\theta$ affects the training of the GenQA model. We consider the GAVA-SDA approach, and the MS-MARCO dataset, and use three different values of $\theta {=} \{0.5, 0.7, 0.9\}$. We follow the same experimental setting described in Section~\ref{ssec:models}, and present results in Table~\ref{tab:results_thresdholds_ablation}. The results suggest a trend of achieving a higher final GenQA accuracy using a higher value of $\theta$. This highlights that the quality of the generated answers (for augmenting) is more important for downstream answer generation than the quantity (Higher $\theta$ will pick ``better-quality" answers, but increase the number of answers getting filtered out). Additionally we plot the GAVA-Score on the development set across training with these different values of $\theta$ in Fig.~\ref{fig:ablation_thresholds}. We observe that the 
GAVA-Score for the model trained with $\theta{=}0.9$ is always higher than the score of the other models across the entire training.

\mypara{Correlation with other Automatic Metrics} We perform a study to analyze how well can other automatic evaluation metrics perform for the task of evaluating answer generation. Specifically, we consider BLEU, BLEURT, and BERTScore. We use the MS-MARCO dataset, and evaluate several different GenQA models trained using our approaches. We evaluate performance using each of these automatic metrics and GAVA; and present the Pearson and Spearman's rank correlation between these metrics and the manual evaluation in Table~\ref{tab:ablation_correlation}. From the table, we observe that GAVA achieves the strongest correlation with human evaluation, highlighting its efficacy to be used as an automatic QA evaluation metric. Other text similarity matching metrics achieve poor correlation with the human annotation of answer correctness.

\begin{table*}[h]
\small
\begin{center}
\resizebox{1.0 \linewidth}{!}{
    \begin{tabular}{rll} 
        \toprule
    1)   & Question        &  How do I get from DC to Alexandria VA? \\
        &  Answer \#1    & Check out the Metrorail system map and WMATA's official trip planner to plot the best route to your destination. \\
        &  Answer \#2    & To get to downtown DC from Alexandria, take the Yellow Line toward Greenbelt or Fort Totten, or take the Blue \\
        &                    & Line toward Largo Town Center or Stadium-Armory.Check out the Metrorail system map and  \\ 
        &  Answer \#3    & WMATA's official trip planner to plot the best route to your destination.o get to downtown DC from Alexandria, take \\
        &                    & the Yellow Line toward Greenbelt or Fort Totten, or take the Blue Line toward Largo  Town Center or Stadium-Armory. \\
        &  Answer \#4    & In addition to the King St-Old Town Metrorail station, Alexandria is serviced by three other stations: Braddock Road, \\
        &                    & Eisenhower Ave, and Van Dorn Street. \\
        &  \textcolor{green}{\textbf{GenQA Answer}}        &  The best way to get to Alexandria from Washington, DC is by Metrorail. \\
        \midrule
    2)  &  Question         & How long should a central air conditioner last? \\
       
        &  Answer \#1    &  10 to 20 years - sometimes longer. \\
        &  Answer \#2    &  You wouldn't run your car without regular maintenance and tune ups and you shouldn't run your a/c that way either \\
        &                   &  - if you want it to last as long as possible. You should have a service tech come out once a year for a tune up. \\
        &  Answer \#3    &  Source(s): 20 years working for a major manufacturer of central heating and air conditioning. \\
         &  \textcolor{green}{\textbf{GenQA Answer}}       & The average life span of a central air conditioner is 10 to 20 years. \\
        \midrule
    3)  &  Question      & How long before a bearded dragon starts to grow \\
        &  Answer \#1 &  When grown, males are larger than females, with notably bigger heads. \\ 
        &  Answer \#2 &  Determining the gender of your bearded dragon before maturity isn't easy. \\
        &  Answer \#3 &  Bearded dragons reach sexual maturity between the ages of 8 months and 18 months. \\
        & \textcolor{green}{\textbf{GenQA Answer}}      &  Bearded dragons reach sexual maturity between the ages of 8 months and 18 months. \\
       \bottomrule
    \end{tabular}}
    \caption{Examples of correctly generated answers using GAVA-DDA approach on the MS-MARCO dataset. Example (1) highlights that the model is correctly able to synthesize a correct answer using the reference answer candidates for the question. Example (2) highlights a case where the GenQA model uses information from a single reference answer, but reformulates it's style using the question to present as an answer. Example (3) highlights a case where the GenQA model is effectively functioning as an answer ranker, as it directly copies the best answer candidate among the references to produce the generated answer. }
    \label{tab:qualitative_analysis_examples_good}
\end{center}
\end{table*}

\begin{table*}[ht]
\small
\begin{center}
\resizebox{1.0 \linewidth}{!}{
    \begin{tabular}{rll} 
        \toprule
     1) & Question        &  Albany MN population ?  \\
        &  Answer \#1  &  Albany, Minnesota, as per 2017 US Census estimate, has a community population of 2,662 people. \\
        & Answer \#2   &  Albany is located in Stearns County, 20 miles west of St. Cloud and 80 miles northwest of Minneapolis/St. Paul \\
        &                 &  on Interstate 94 (I-94). Albany has direct access to State Highway 238, which originates in Albany.  \\
        &  \textcolor{red}{\textbf{GenQA Answer}}       &  The population was 2,662 at the 2010 census. \\
       \midrule
     2) & Question        &  Which layer is responsible for the earth's magnetic field? \\
        & Answer \#1   & Best Answer: The Earth's magnetic field is produced by convective currents in the outer core. \\
        & Answer \#2   & The (presumably) molten iron core. \\
        & Answer \#3   & The outer core is liquid iron. \\
        & \textcolor{red}{\textbf{GenQA Answer}}        &  The outer core is liquid iron. \\
        \bottomrule
    \end{tabular}}
    \caption{Examples of incorrectly generated answers using GAVA-DDA approach on the MS-MARCO dataset. Example (1) highlights a case of hallucination during generation where the model introduces an incorrect year in the generated answer, even when it is not present in any of the input reference candidates. Example (2) highlights a failure case of GenQA where the model is unable to synthesize a good answer due to lacking evidence in the retrieved reference candidates.}
    \label{tab:qualitative_analysis_examples_bad}
\end{center}
\end{table*}

\subsection{Qualitative results}
\label{ssec:qualitativeresults}


We perform a qualitative analysis highlighting anecdotal examples to study success and failure cases of our answer-generation approach. Specifically, we pick the MS-MARCO dataset and the GAVA-DDA approach, and present both success and failure cases of answer generation to gain insights into the strengths and limitations of our approach. 

Table~\ref{tab:qualitative_analysis_examples_good} shows instances where GAVA-DDA successfully generates accurate answers. These examples highlight various sub-tasks that the model implicitly performs. Firstly, the model demonstrates its ability to synthesize information from multiple answer candidates. For example, for the question \textit{"How do I get from DC to Alexandria VA?"}, the model correctly synthesized information from each of the reference answer candidates into the generated answer about the \textit{Metrorail service} connecting the two locations. Second, the model exhibits reasoning capabilities highlighting identification of the correct reference answer candidate, along with improving it's style suitability for answering the input question. This is observed in the second example with the question \textit{"How long should a central air conditioner last?"}, where the model identified the first reference \textit{"10 to 20 years - sometimes longer"} to contain the most relevant information for answering the question. At times, the model acts as an answer sentence selection (AS2) model that simply re-ranks (without any modification) and generates one of the reference answer candidates.

Table~\ref{tab:qualitative_analysis_examples_bad} presents some examples where the model hallucinates and produces incorrect answers. This is highlighted in the question about Albany Minnesota's Population, where the model hallucinates and introduces an incorrect year in the generated answer, even when it is not present in any of the input reference candidates. Additionally, at times, the model may be unable to synthesize a good answer due to lacking evidence in the retrieved reference candidates. This is highlighted in the question about the earth's magnetic field. This limitation emphasizes the importance of reliable and accurate answer candidates for grounding the answer generation from the model. 

\section{Conclusion}
\label{sec:conclusion}
\vspace{-0.3em}

In this paper we propose a novel training paradigm of learning answer generation systems (GenQA) using supervision from automatic QA evaluation metrics based on transformer encoders. We propose three strategies: augmenting the training corpora with high GAVA-scoring generated answers for training the GenQA model (either statically once before training, or dynamically at every training epoch); and using the GAVA score for weighting the loss during the learning of the GenQA model. We perform empirical evaluation on two academic and one industrial dataset and show that our approaches outperform the baseline with both human annotations and automatic QA evaluation metrics (GAVA score). For future work, we plan to investigate how automatic QA evaluator based preferences align with human-annotated preferences for training larger LMs via reinforcement learning~\cite{lambert2022illustrating}. This would involve using GAVA as the RLHF reward model.

\vspace{-0.5em}
\section*{Limitations}
\label{sec:limitations}
\vspace{-1em}
The main limitation of our methodology is that the training of Generative Question Answering models requires the usage of large GPU resources, which may not be easily available to all researchers. Regarding the performance of the model and quality of the generated answers, our approach can be affected by possible bias induced by the evaluation system we are using. For the experiments in this paper, we only consider datasets from the English language, however, we conjecture that our techniques should work similarly for languages with a similar morphology. Automatic QA evaluation systems do not achieve perfect correlation with human annotations, which indicates a gap with respect to human evaluation. For safety critical applications, human evaluation of generated answers still remains the best
means for evaluation.
 
\bibliography{anthology,custom}
\bibliographystyle{acl_natbib}

\clearpage
\appendix

\section*{Appendix}

\section{Implementation Details}
\label{apx:implementationdetails}

\subsection{Computational Setting}
We train our models on a machine with $8$ NVIDIA A100 with 40Gb of VRAM and $1.1$Tb of RAM. Our framework is based on Pytorch \cite{pytorch} and hugging face \cite{lhoest-etal-2021-datasets, wolf-etal-2020-transformers}.  

\subsection{GAVA training}

The multi-reference GAVA models are structurally closely related with the multi-sentence answer selection models proposed in ~\cite{multi-sentence2022}. We train two different multi-reference GAVA models starting from a DeBERTaV3-Large model \cite{he2021debertav3} on two different datasets: WQA for the experiments on public dataset, and IQAD for the industrial scenario. For both the settings we use Adam \cite{Kingma2015AdamAM} with a learning rate of $1e-06$, a batch size of $32$ and $fp32$ for $20$ epochs. We shuffle the training set at the beginning of each epoch and we evaluate our model $4$ times on the development set considering different performance measure. At the end of the training, we select the best checkpoint maximizing the area-under-the-curve (AUROC) on the development set.


\subsection{GenQA based models training}

We train our approaches on WQA, MS-MARCO and IQAD starting from a T5-Large model pretrained using WS on MS-MARCO \cite{gabburo2022}. To train the models we use Adam as optimizer with lr=$5e-06$, $fp32$, and batch size of $32$ shuffling the training set at the beginning of each epoch. For MS-MARCO, we train the model for $15$ epochs while for WQA and IQAD, we train the model for $30$ epochs. We select the best checkpoint in term of GAVA-Score computed on the development set. We adopt an early stopping criterion stopping the training when the model does not improve for  $3$ epochs.

\end{document}